\documentclass[11pt]{article}
\usepackage{eamt26}
\usepackage{times}
\usepackage{url}
\usepackage{latexsym}
\usepackage[small,bf]{caption} 
\usepackage{graphicx}
\usepackage{multirow}
\usepackage{booktabs} 
\usepackage[T1]{fontenc}
\usepackage[utf8]{inputenc}
\usepackage{microtype}
\usepackage{enumitem}
\usepackage{inconsolata}
\usepackage{graphicx}
\usepackage{multirow}
\usepackage{etoolbox}
\AtBeginEnvironment{table}{\setlength{\belowcaptionskip}{-5pt}\setlength{\abovetopsep}{0pt}}

\title{Towards Visually-Guided Movie Subtitle Translation for Indic Languages}

\author{Tarun Chintada\thanks{~~Equal contribution.}, Kshetrimayum Boynao Singh\footnotemark[1], Asif Ekbal \\
  Department of Computer Science and Engineering \\
  Indian Institute of Technology Patna, India \\
  \texttt{\{tarunchintada1, boynfrancis, asif.ekbal\}@gmail.com}}

\begin{document}
\maketitle

\begin{abstract}
Movie subtitle translation is inherently multimodal, yet text-only systems often miss visual cues needed to convey emotion, action, and social nuance, especially for low-resource Indic languages (English to Hindi, Bengali, Telugu, Tamil and Kannada). We present a case study on five full-length films and compare two lightweight visual grounding strategies: structured attribute summaries from a 5-minute sliding window and free-text summaries of inter-subtitle visual gaps. Our analysis shows that temporal misalignment between subtitles and frames is a major obstacle in long-form video, often rendering indiscriminate visual grounding ineffective. However, oracle selective\footnote{Oracle selective refers to a specialized, often theoretical, method used to achieve the absolute best outcome} grounding, which replaces only the lowest-quality 20-30\% of baseline segments with visual-enhanced outputs, consistently improves COMET over the text-only baseline while requiring far less visual processing. Among the two approaches, coarse attribute-based visual context summarization is more robust, capturing scene-level emotion and contextual subtle cues that text alone often misses.
\end{abstract}

\section{Introduction}
The global demand for movie subtitle translation has grown exponentially with the rise of streaming platforms and international releases. Subtitles must convey meaning under strict spatial and temporal constraints, often compressing conversational speech, idiomatic expressions, and cultural references into short, timed segments. For low‑resource Indian languages characterised by rich morphology \cite{singh-etal-2025-evaluating}, diglossia, and sparse parallel corpora, these challenges are magnified, and text‑only machine translation (MT) systems frequently produce translations that are either literal or contextually inadequate \cite{singh-etal-2025-evaluation,artetxe-etal-2020-cross-b}.

\begin{figure*}[t]
    \centering
    \includegraphics[width=\textwidth]{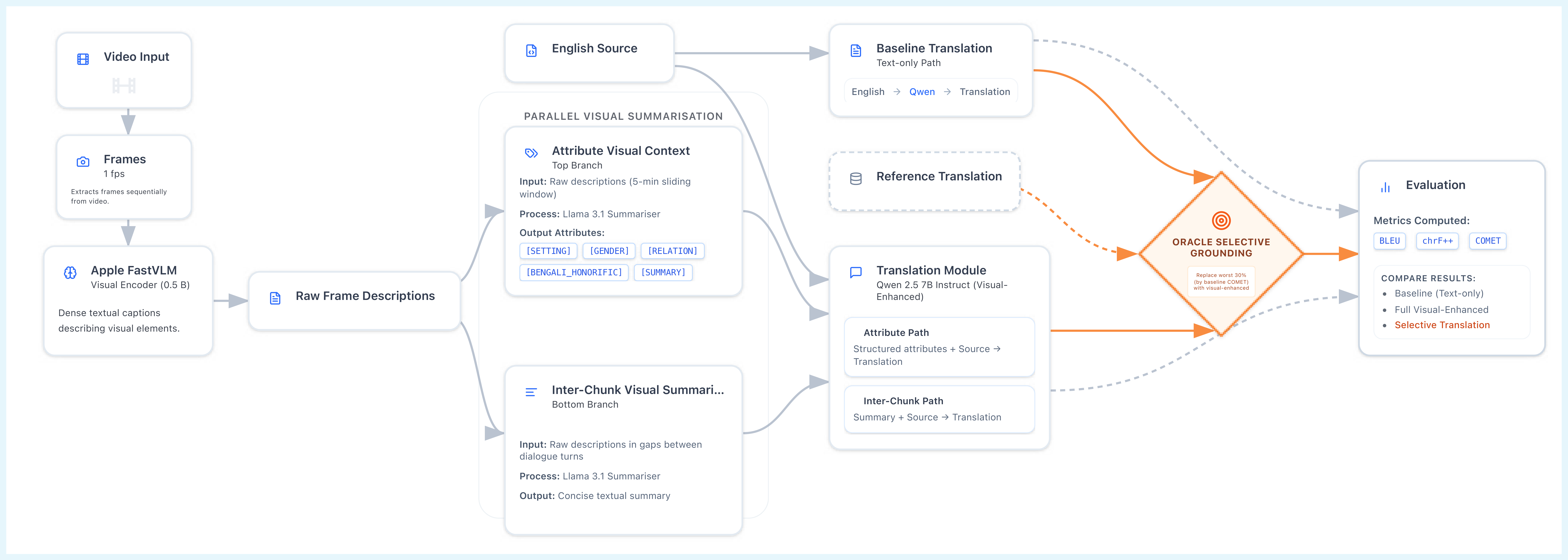}
    \caption{Architecture of the multimodal subtitle translation pipeline}
    \label{fig:Archi}
\end{figure*}

Films are inherently multimodal: meaning is distributed across dialogue, visual scenes, character actions, and emotional cues. In principle, incorporating visual context could disambiguate references, resolve honorifics, and ground translations in the on‑screen situation~\cite{elliott2016multi30k}. Yet, unlike traditional multimodal machine translation (MT) tasks (e.g., translating image captions), movie subtitle translation presents two distinctive difficulties.

\textit{First, most subtitle segments do not rely on visual information.} The majority of dialogues are conversational and can be translated accurately from text alone. Visual grounding is beneficial only in a minority of cases-action cues~\cite{10.1145/3748311,kumar-etal-2025-vision}, emotion‑driven~\cite{shen-etal-2025-coe} exchanges, or references to visible objects. For instance, translating the line \textit{“He’s coming!”} requires knowing whether the threat is a person, animal, or vehicle; such information is visually available. In contrast, a typical exchange like \textit{“How was your day?”} gains little from the accompanying visuals. This asymmetry makes indiscriminate application of computationally expensive visual processing inefficient and often unnecessary.

\textit{Second, visual and textual streams in long‑form movies are often misaligned.} Subtitles are generated independently of video frames, and cumulative temporal drift can cause a substantial fraction of subtitles to be paired with irrelevant or misleading visuals~\cite{wang-zhao-2024-tram}. Over a 180‑minute film, drift as small as one second per hour accumulates to a three‑minute mismatch, affecting a notable portion of subtitle segments. When visual context is misaligned, it ceases to be helpful and can actively degrade translation quality~\cite{appicharla-etal-2024-case}, a phenomenon rarely discussed in the multimodal MT literature~\cite{radinger2025subtitling}.

Motivated by these practical realities, we conduct a case study that systematically compares two summarization‑based strategies for integrating visual context into subtitle translation for five Indian languages: Hindi, Bengali, Telugu, Kannada, and Tamil. These languages represent a range of morphological complexity and cultural nuances, making them ideal for studying multimodal subtitle translation in low‑resource settings~\cite{eberhard2025ethnologue}. The two strategies are:
\begin{enumerate}[noitemsep]
    \item \textbf{Attribute Visual Context (Attr‑VC)}: Aggregates a 5‑minute sliding window of raw visual descriptions and summarizes them using Llama 3.1 into structured attributes (e.g., setting, gender, honorifics, emotional intent). \\
    \item \textbf{Inter‑Chunk Visual Summarization (Inter-VS)}: Summarizes the visual content occurring between dialogue turns (the gaps) into a free‑text description using Llama 3.1.
\end{enumerate}

Using subtitles from five full‑length movies spanning diverse genres, we evaluate these methods under realistic conditions. A central finding is that applying visual context indiscriminately often degrades performance due to temporal misalignment. However, an \emph{oracle selective grounding} that replaces the worst 20-30\% of baseline segments (by baseline COMET) with visual‑enhanced translations consistently improves semantic adequacy (COMET) over the text‑only baseline, recovering most of the gain while using only a fraction of the visual processing. Coarse attribute‑based summarization proves particularly robust, capturing emotional tone and scene‑level cues that text alone cannot convey. The key attribute of this work are:
\begin{enumerate}[noitemsep]
    \item \textbf{A comparative case study} of two visual summarization strategies for low‑resource subtitle translation.
    \item \textbf{Identification and quantification} of temporal misalignment as a major practical obstacle in long‑form multimodal MT.
    \item \textbf{Empirical evidence} that coarse attribute summarization is resilient to drift and that selective grounding can recover most of the gain.
\end{enumerate}

\begin{table*}[t]
\centering
\small
\setlength{\tabcolsep}{5pt}
\resizebox{\textwidth}{!}{
\begin{tabular}{l l c c c}
\hline
\textbf{Movie} & \textbf{Genre} & \textbf{Duration} & \textbf{Total Frames} & \textbf{Extracted Frames} \\
\hline
Titanic (1997)      & Romance, Drama, Epic               & 3:14:54 & 280,305 & 11,694 \\
Skyfall (2012)      & Action, Adventure, Spy Thriller    & 2:23:10 & 205,952 & 8,589 \\
Oppenheimer (2023)  & Biographical, History, Drama       & 3:00:22 & 259,472 & 10,822 \\
Spider‑Man 2 (2004) & Superhero, Action, Sci‑Fi          & 2:15:48 & 195,360 & 8,148 \\
Avatar 2 (2022)     & Sci‑Fi, Action, Adventure          & 3:12:38 & 277,117 & 11,558 \\
\hline
\end{tabular}
}
\caption{Visual data statistics for the five movies. Extracted frames are sampled within the time span of each subtitle segment.}
\label{tab:movie_visual_stats}
\end{table*}

\section{Dataset Preparation and Resources}
To evaluate multimodal subtitle translation in realistic settings, we curate a dataset derived from five commercially released movies selected to ensure diversity in genre, narrative style, and visual complexity: \textit{Titanic} (1997), \textit{Skyfall} (2012), \textit{Oppenheimer} (2023), \textit{Spider‑Man 2} (2004), and \textit{Avatar 2} (2022). These films span romance, action, historical drama, science fiction, and superhero genres, providing a wide range of dialogue types and visually grounded scenes.

\subsection{Movies and Visual Data}
For each movie, we extract video frames at 24-fps and align them with subtitle timestamps. Table~\ref{tab:movie_visual_stats} summarizes the visual data, including total duration, total frames, and the number of frames extracted within subtitle time spans (i.e., frames that fall within the time window of a subtitle segment). The extracted frames serve as the visual input for our multimodal methods.

\subsection{Subtitle Corpora}
We extract subtitles from publicly available sources\footnote{http://subtitlecat.com} and temporally align them with the corresponding video segments. All subtitle pairs are preprocessed to remove noise, normalize punctuation, and filter excessively long or short segments. Table~\ref{tab:subtitle_stats} provides detailed statistics for each movie, including the number of subtitle pairs (English source and target language), average English source length in words, and average English source length in characters. Parallel subtitles are available for Hindi (all movies except Avatar 2), Bengali and Telugu (all movies), Kannada (all movies except Spider‑Man 2), and Tamil (all movies except Titanic). This selection ensures coverage of linguistically diverse Indian languages while respecting the availability of high‑quality parallel subtitles. All movies were legally purchased as DVDs. Frame extraction for research constitutes fair use and follows standard practice in video‑language benchmarks.

\begin{table}[t]
\centering
\small
\setlength{\tabcolsep}{4pt}
\resizebox{\columnwidth}{!}{
\begin{tabular}{l c c c}
\hline
\textbf{Movie} & \textbf{Total Pairs} & \textbf{Avg Words} & \textbf{Avg Chars} \\
\hline
Titanic      & 1,991 & 5.64 & 29.57 \\
Skyfall      & 1,140 & 5.70 & 29.74 \\
Oppenheimer  & 3,519 & 5.68 & 31.77 \\
Spider‑Man 2 & 1,049 & 5.83 & 30.64 \\
Avatar 2     & 1,444 & 6.44 & 32.86 \\
\hline
\textbf{Total} & \textbf{9,143} & \textbf{5.86} & \textbf{30.92} \\
\hline
\end{tabular}
}
\caption{Subtitle statistics per movie. Total pairs represent the number of English subtitle segments.}
\label{tab:subtitle_stats}
\end{table}

\subsection{Data Release}
To foster reproducibility and further research, we will release the curated movie-subtitle-visual alignment data for all five languages under a fair‑use educational/research license. The release includes the English source, reference translations, and extracted visual descriptions. The code and instructions for reproducing the experiments will also be made publicly available.

\section{Methodology}
Our methodology is designed for real-world applicability: all models are used off-the-shelf in a zero-shot setting, no fine-tuning or training is performed, and the pipeline is fully reproducible. We use Qwen-2.5-7B-Instruct~\cite{qwen2025qwen25technicalreport} as the translation model, Llama-3.1-8B-Instruct~\cite{llama3} for summarization~\cite{datta-etal-2025-findings}, and Apple FastVLM-0.5B~\cite{vasu2025fastvlmefficientvisionencoding} for visual description extraction. The pipeline is depicted in Figure~\ref{fig:Archi}. For the text‑only baseline, Qwen is prompted with the English source only. This yields the text‑only translation against which visual‑enhanced methods are compared.

\subsection{Visual Context Generation}
From each movie, we sample frames at 1-fps and obtain raw textual descriptions using FastVLM-0.5B. These descriptions are then summarized by Llama-3.1 into two distinct forms:

\paragraph{Attribute Visual Context (Attr‑VC)} A 5‑minute sliding window (centered on the subtitle start) is aggregated and summarized into structured attributes: \texttt{[SETTING]}, \texttt{[GENDER]}, \texttt{[RELATION]}, \texttt{[HONORIFIC]}, and \texttt{[SUMMARY]}. This yields a coarse, high‑level scene description.

\paragraph{Inter‑Chunk Visual Summarization (Inter-VS)} The raw descriptions that fall between the end of the previous subtitle and the start of the current subtitle (the visual gap) are summarized into a free‑text description. This captures visual events that occur between dialogue turns. The full prompts used for these summarization tasks are provided in Table~\ref{tab:Summarisation_prompts}.
Both the summaries are concatenated with the English source using the same prompt template, which instructs the model to ground its translation in the visual context. 

\subsection{Oracle Selective Grounding}
To estimate the upper bound of selective visual grounding, we compute per‑segment COMET scores for the baseline translations against the reference. We then replace the worst \(k\%\) of segments (by baseline COMET) with the corresponding visual‑enhanced translation (from either Attr‑VC or Inter-VS). We experiment with \(k = 20\%\) and \(30\%\). This oracle analysis shows the potential improvement if one could perfectly identify low‑quality baseline segments; it does not require any training and represents an upper bound for practical quality‑estimation systems.


\section{Evaluation Results }
\subsection{Evaluation Setup}
We evaluate on the full curated test set (all aligned subtitle segments) using corpus‑level BLEU~\cite{papineni-etal-2002-bleu}, chrF++~\cite{popovic-2015-chrf}, and COMET~\cite{rei-etal-2020-comet}. 

\subsection{Results and Analysis} We compare the two visual summarisation strategies \emph{Attribute Visual Context (Attr‑VC)} and \emph{Inter‑Chunk Visual Summarization (Inter-VS)} against a text‑only baseline. We perform experiments with five movies in five Indian languages (Hindi, Bengali, Telugu, Tamil, Kannada) using Qwen-2.5-7B.

Results for full per‑movie, per‑language are shown in Table~\ref{tab:combined_results}. Table~\ref{tab:lang_summary} summarizes the language‑wise COMET improvements.

\begin{table*}[t]
\centering
\small
\resizebox{\textwidth}{!}{%
\begin{tabular}{l l c c c|c c c|c c c|c c c|c c c}
\toprule
\multirow{2}{*}{Movie} & \multirow{2}{*}{Lang} & \multicolumn{3}{c}{Baseline} & \multicolumn{6}{c}{5‑Minute Slide Visual Attribute} & \multicolumn{6}{c}{Inter‑Chunk Visual Summarisation} \\
\cmidrule(lr){3-5} \cmidrule(lr){6-11} \cmidrule(lr){12-17} 
 & & BLEU & chrF++ & COMET & \multicolumn{3}{c}{Visual-Enhanced} & \multicolumn{3}{c}{Oracle Selective} & \multicolumn{3}{c}{Visual-Enhanced} & \multicolumn{3}{c}{Oracle Selective} \\
\cmidrule(lr){6-8} \cmidrule(lr){9-11} \cmidrule(lr){12-14} \cmidrule(lr){15-17}
 & & & & & BLEU & chrF++ & COMET & BLEU & chrF++ & COMET & BLEU & chrF++ & COMET & BLEU & chrF++ & COMET \\
\midrule
Avatar & Ben  & 5.68 & 28.71 & 0.6298 & 6.95 & 27.95 & \textbf{0.7014} & 6.92 & \textbf{29.98} & \textbf{0.6829} & \textbf{8.10} & 28.24 & 0.7137 & 6.91 & 29.80 & 0.6865 \\
Avatar & Tel  & 4.30 & 19.66 & 0.5257 & 3.28 & 18.23 & 0.5154 & 4.38 & \textbf{19.79} & 0.5390 & \textbf{3.67} & 18.32 & 0.5153 & \textbf{4.67} & 19.85 & 0.5390 \\
Avatar & Tam  & 3.85 & 23.49 & 0.5352 & 4.08 & 22.94 & 0.5545 & 4.24 & 24.36 & 0.5580 & \textbf{4.57} & \textbf{23.62} & \textbf{0.5613} & \textbf{4.33} & \textbf{24.50} & \textbf{0.5639} \\
Avatar & Kan  & 3.50 & 18.94 & 0.4857 & 2.34 & 15.20 & 0.4582 & 3.39 & 18.65 & 0.4933 & \textbf{2.23} & \textbf{15.28} & \textbf{0.4612} & \textbf{3.33} & 18.37 & \textbf{0.4946} \\
\midrule
Oppenh. & Ben  & 8.05 & 29.38 & 0.7026 & 5.47 & 25.09 & 0.6735 & 8.03 & \textbf{29.41} & 0.7248 & \textbf{6.37} & 26.26 & 0.6858 & \textbf{8.08} & 29.50 & 0.7237 \\
Oppenh. & Hin  & 11.76 & 31.64 & 0.6467 & 8.62 & 27.28 & 0.5972 & 11.83 & 31.74 & 0.6642 & \textbf{9.62} & \textbf{28.72} & \textbf{0.6297} & \textbf{11.86} & \textbf{32.06} & \textbf{0.6690} \\
Oppenh. & Tel  & 4.04 & 18.86 & 0.5475 & 3.29 & 17.77 & 0.5387 & 3.89 & \textbf{19.13} & \textbf{0.5654} & \textbf{3.53} & 18.05 & 0.5379 & \textbf{3.96} & 19.21 & 0.5647 \\
Oppenh. & Tam  & 3.15 & 20.60 & 0.5366 & 3.15 & 21.37 & 0.5654 & 3.36 & 21.85 & 0.5630 & \textbf{3.47} & \textbf{21.63} & \textbf{0.5690} & \textbf{3.66} & \textbf{22.18} & \textbf{0.5715} \\
Oppenh. & Kan  & 2.95 & 16.81 & 0.4938 & 2.23 & 14.63 & 0.4740 & 2.96 & 17.20 & 0.5066 & \textbf{2.30} & 14.25 & 0.4735 & 2.93 & 17.05 & \textbf{0.5090} \\
\midrule
Skyfall & Ben  & 6.31 & 27.30 & 0.6914 & 4.10 & 23.51 & 0.6588 & \textbf{6.04} & 27.39 & \textbf{0.7098} & \textbf{4.55} & \textbf{23.68} & 0.6612 & 5.93 & 27.14 & 0.7056 \\
Skyfall & Hin  & 6.31 & 25.65 & 0.6026 & 5.74 & 25.28 & 0.5882 & \textbf{6.53} & \textbf{26.68} & \textbf{0.6258} & \textbf{6.41} & 25.86 & 0.6098 & 6.65 & 26.47 & 0.6245 \\
Skyfall & Tel  & 2.47 & 17.68 & 0.5288 & 2.13 & 16.66 & 0.5248 & \textbf{2.24} & \textbf{18.00} & \textbf{0.5478} & 1.41 & 16.84 & 0.5157 & 2.16 & 17.86 & 0.5454 \\
Skyfall & Tam  & 2.33 & 21.09 & 0.5350 & 2.22 & 21.11 & 0.5581 & 2.59 & 21.78 & 0.5581 & \textbf{1.83} & 20.89 & \textbf{0.5595} & \textbf{2.66} & \textbf{22.02} & \textbf{0.5639} \\
Skyfall & Kan  & 1.59 & 16.88 & 0.4920 & 1.76 & 14.38 & 0.4668 & 1.59 & 16.90 & 0.5013 & \textbf{1.54} & \textbf{14.36} & \textbf{0.4682} & 1.58 & 16.80 & \textbf{0.5038} \\
\midrule
Spider2 & Ben  & 9.58 & 26.81 & 0.7190 & 6.69 & 24.44 & 0.6902 & 9.10 & 26.97 & 0.7359 & \textbf{8.55} & \textbf{25.77} & \textbf{0.7021} & \textbf{9.47} & \textbf{27.18} & \textbf{0.7350} \\
Spider2 & Hin  & 12.33 & 29.15 & 0.6459 & 10.13 & 27.71 & 0.6286 & 12.61 & \textbf{30.20} & 0.6746 & \textbf{12.19} & 29.17 & \textbf{0.6532} & \textbf{12.93} & 30.32 & \textbf{0.6786} \\
Spider2 & Tel  & 5.22 & 18.47 & 0.5407 & 3.57 & 17.69 & 0.5349 & 5.04 & 18.84 & 0.5567 & 3.40 & 17.73 & 0.5342 & \textbf{5.04} & \textbf{18.74} & \textbf{0.5569} \\
Spider2 & Tam  & 4.12 & 21.65 & 0.5448 & 3.27 & 21.39 & 0.5601 & 4.29 & 22.73 & 0.5684 & \textbf{4.01} & \textbf{22.09} & \textbf{0.5694} & \textbf{4.32} & \textbf{22.83} & \textbf{0.5761} \\
\midrule
Titanic & Ben  & 9.59 & 25.87 & 0.6960 & 7.04 & 22.97 & 0.6616 & 9.55 & 26.11 & 0.7130 & \textbf{8.65} & \textbf{24.97} & \textbf{0.6849} & \textbf{9.82} & \textbf{26.41} & \textbf{0.7150} \\
Titanic & Hin  & 11.98 & 26.59 & 0.6152 & 9.12 & 24.45 & 0.5711 & 11.92 & 27.12 & 0.6321 & \textbf{12.29} & \textbf{26.90} & \textbf{0.6176} & \textbf{12.66} & \textbf{27.62} & \textbf{0.6367} \\
Titanic & Tel  & 5.03 & 17.82 & 0.5350 & 3.85 & 16.99 & 0.5211 & 4.92 & \textbf{18.01} & 0.5481 & \textbf{4.32} & 17.52 & 0.5296 & \textbf{5.11} & 18.21 & 0.5494 \\
Titanic & Kan  & 4.95 & 17.16 & 0.4950 & 3.11 & 14.04 & 0.4670 & 4.73 & 16.97 & 0.5037 & \textbf{3.14} & \textbf{14.45} & \textbf{0.4665} & \textbf{4.86} & \textbf{17.03} & \textbf{0.5049} \\
\bottomrule
\end{tabular}
}
\caption{Comparison of two visual summarization strategies. \emph{5‑Minute Slide Visual Attribute} aggregates a 5‑minute sliding window into structured attributes (setting, gender, honorifics, emotion); \emph{Inter‑Chunk Visual Summarization} summarizes the visual content between dialogue turns. For each method, we report \emph{Visual-Enhanced} (using the full visual context for all segments) and \emph{Oracle Selective} (replacing the worst 30\% of baseline segments by baseline COMET with the visual‑enhanced translation). This oracle shows the upper bound of selective grounding. Metrics are corpus‑level BLEU, chrF++, and COMET. \textbf{Bold} indicates the higher score between the two methods for the same condition (Visual-Enhanced or Oracle Selective).}
\label{tab:combined_results}
\end{table*}

\begin{table}[t]
\centering
\small
\begin{tabular}{l c c c c}
\toprule
\multirow{2}{*}{Language} & \multicolumn{2}{c}{Attr-VC} & \multicolumn{2}{c}{Inter-VS} \\
\cmidrule(lr){2-3} \cmidrule(lr){4-5}
 & Full & Sel30 & Full & Sel30 \\
\midrule
Hindi     & -5.0\% & +3.4\% & 0.0\%  & +3.9\% \\
Bengali   & -1.6\% & +3.7\% & +0.3\% & +3.7\% \\
Telugu    & -1.6\% & +3.0\% & -1.7\% & +2.9\% \\
Tamil     & +4.0\% & +5.9\% & +5.0\% & +5.8\% \\
Kannada   & -5.1\% & +2.3\% & -4.9\% & +2.4\% \\
\bottomrule
\end{tabular}
\caption{Language-wise average COMET improvement ($\Delta$) over baseline for each method and condition. Positive values indicate improvement. Oracle Selective replaces the worst 30\% of baseline segments by baseline COMET.}
\label{tab:lang_summary}
\end{table}

\begin{figure}[t]
\centering
\includegraphics[width=\columnwidth]{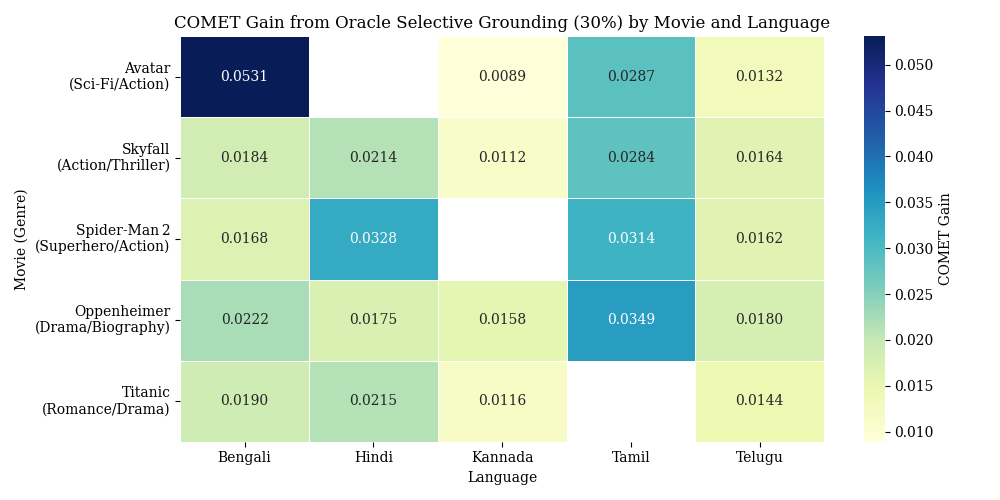}
\caption{COMET gain from Oracle Selective Grounding (30\%) by movie and language. Gain is computed as the difference between the oracle selective translation (replacing the worst 30\% of baseline segments by baseline COMET) and the text‑only baseline, using the better of the two visual summarisation methods for each pair. Movie names are followed by their genre in parentheses. Darker shades indicate larger gains.}
\label{fig:gain_heatmap}
\end{figure}

\subsection{Overall Observations}
Both summarisation methods show that applying visual context indiscriminately often degrades performance compared to the text‑only baseline. For example, in several movie-language pairs (e.g., Oppenheimer Hindi, Skyfall Bengali),  full VT COMET is lower than the baseline. This is directly attributable to temporal misalignment: when visual frames do not match the spoken dialogue, the model is misled. However, oracle selective grounding consistently improves COMET over the baseline in almost all cases, recovering most of the potential gain while using only 30\% of the visual processing. Figure~\ref{fig:gain_heatmap} visualises the per‑movie, per‑language COMET gain from oracle selective grounding, highlighting that action‑rich movies (e.g., Skyfall) show larger improvements. Human evaluation with a small set (30 examples, each for Telugu and Hindi) confirmed that selective grounding with oracle significantly improves adequacy: average score increased from 2.9 (baseline) to 4.1 (selective) on a 1-5 scale.
 
\subsection{Comparison of Summarization Methods}
Attr‑VC, which aggregates a 5‑minute window into coarse attributes, proves more robust to drift than Inter-VS. In many cases (e.g., Avatar Bengali, Oppenheimer Telugu, Titanic Hindi), its selective 30\% COMET surpasses the full VT of Inter-VS. The attribute‑based representation, by ignoring precise timing, effectively filters out irrelevant frames. Inter-VS, while capturing finer‑grained visual events, is more sensitive to misalignment; its full VT often underperforms the baseline, but selective grounding still yields gains.

\begin{figure}[t]
\centering
\includegraphics[width=\columnwidth]{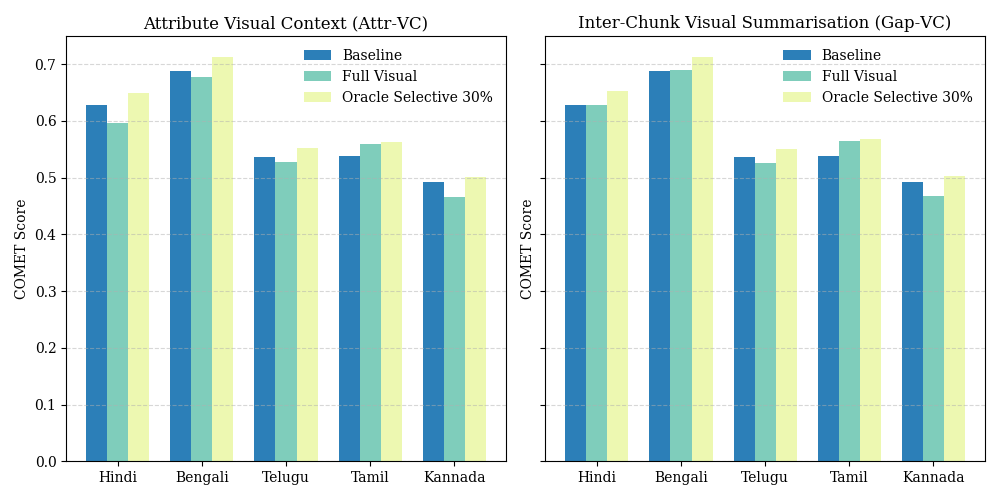}
\caption{Language‑wise COMET scores for the two visual summarization methods. For each method, bars show the baseline (text‑only), full visual‑enhanced translation, and oracle selective grounding (replacing the worst 30\% of baseline segments by baseline COMET). The oracle selective consistently improves COMET over the baseline across all languages, with relative gains of 2-5\%.}
\label{fig:comet_lang}
\end{figure}

\subsection{Language‑Wise Trends}
Table~\ref{tab:lang_summary} aggregates COMET improvement over the baseline for each language. For Attr‑VC selective, all languages show positive gains (range +2.3\% to +5.9\%). The gains are larger for morphologically rich languages (Bengali, Tamil, Kannada) where visual cues (e.g., honorifics, emotional tone) help resolve pragmatic ambiguities. Inter-VS selective also yields consistent improvements, though slightly lower for some languages. The COMET improvements across languages are further illustrated in Figure~\ref{fig:comet_lang}.

\subsection{Summary of Key Findings}
Our case study yields three actionable insights:
\begin{enumerate}
    \item \textbf{Coarse attribute‑based summarization is robust to temporal drift} By aggregating visual information over a 5‑minute window into structured attributes, Attr‑VC achieves pragmatic gains without being misled by misaligned frames.
    \item \textbf{Selective visual grounding can recover most of the gain} An oracle that replaces only the worst 20-30\% of baseline segments with visual‑enhanced translations consistently improves COMET over baseline, using a fraction of the visual processing.
    \item \textbf{Alignment quality often outweighs architectural complexity} Fine‑grained summarization methods that rely on precise temporal alignment are fragile. For real‑world deployment, drift‑tolerant architectures are preferred.

\end{enumerate}

\section{Discussion}
Our case study reveals a central tension: while visual context can provide critical grounding for a small subset of segments, it is irrelevant or even harmful for the majority. This asymmetry, combined with temporal misalignment, shapes the relative performance of the two summarization strategies and the effectiveness of selective grounding.

\subsection{When Visual Context Helps and When It Does Not}
Only a minority of subtitle segments truly depend on visual information. Action cues (“He’s coming!”), emotion‑driven exchanges (“I’m so sorry”), and references to on‑screen objects (“That one”) require visual grounding to resolve ambiguity. For the vast majority of conversational dialogue, text alone is sufficient, and adding visual context adds no benefit. In our dataset, we estimate that fewer than 15\% of subtitles are visually grounded in this sense. This explains why even the best‑performing method oracle selective grounding-achieves only a modest overall COMET gain (2-5\%) while replacing only 20-30\% of segments.

\subsection{Why Attribute Summarization Outperforms Gap Summarization?}
Attr‑VC aggregates a 5‑minute sliding window into high‑level attributes. This coarse representation is rarely misleading for neutral segments and provides valuable pragmatic context for the minority that need it. Moreover, it is inherently robust to misalignment because it aggregates over longer time windows, effectively ignoring irrelevant frames. Inter-VS summarizes the visual content between dialogue turns. This finer‑grained representation captures visual events that may be directly relevant, but it is more sensitive to drift. When visual frames are misaligned, Inter-VS can introduce misleading information, causing its full visual‑enhanced translations to sometimes underperform the baseline. However, when applied selectively (only to the worst baseline segments), Inter-VS still yields substantial gains on the replaced set.

\subsection{The Role of Oracle Selective Grounding}
Our oracle analysis replaces the worst 20-30\% of baseline segments (by baseline COMET) with the corresponding visual‑enhanced translation. This shows the upper bound of what could be achieved with a perfect quality‑estimation model. For both methods, selective grounding consistently lifts COMET above the baseline, often matching or even exceeding the full visual‑enhanced performance of the other method.

\subsection{Insights for Low‑Resource Indian Languages}
For morphologically rich languages such as Bengali and Kannada, the small subset of visually grounded segments often includes honorifics, implicit referents, or emotional tone areas, where text‑only models are weakest. Coarse attribute summarisation helps in these cases without harming the rest, yielding clear COMET gains in the selective setting (e.g., +3.7\% for Bengali, +2.3\% for Kannada). This suggests that for low‑resource settings, investing in reliable, drift‑tolerant visual abstractions is more practical than pursuing fine‑grained fusion or high‑frequency visual processing.

\subsection{Practical Implications for Movie Localisation}
Our findings lead to three actionable recommendations for deploying multimodal subtitle translation:
\begin{enumerate}
    \item \textbf{Be selective:} Not every subtitle needs visual context. An oracle study shows that replacing only the worst 20-30\% of baseline segments can recover most of the gain. A practical system would use a quality‑estimation model (e.g., based on sentence length, emotion words, or visual confidence) to trigger visual enhancement only when needed.
    \item \textbf{Prefer robust architectures:} Attribute‑based summarization (5‑minute sliding window condensed into structured cues) offers a lightweight, drift‑tolerant solution suitable for production pipelines.
    \item \textbf{Fix alignment before using fine‑grained context:} If a more detailed visual context is desired (e.g., gap‑based summarization), pre‑processing steps such as dynamic time warping (DTW) or audio‑visual synchronisation are essential to avoid performance degradation.
\end{enumerate}

\section{Conclusion}
In this paper, we compare two summarization strategies for integrating visual context into subtitle translation for five low‑resource Indian languages. We find that temporal misalignment- a common real‑world issue- causes full visual‑enhanced translation to often underperform the text‑only baseline. However, an oracle selective grounding that replaces only the worst 20-30\% of baseline segments with visual‑enhanced translations consistently improves semantic adequacy (COMET) across all the languages, recovering most of the potential gain while using a fraction of the visual processing. Coarse attribute‑based summarization proves particularly robust to drift, capturing emotional tone and scene‑level cues that text alone cannot convey. Our results underscore that alignment quality often outweighs architectural complexity and that selective visual grounding offers a practical path to efficient, deployable multimodal subtitle translation.

\section*{Limitations}
Our study is limited to five movies and five languages; the proportion of visually grounded segments may vary by genre and across different types of audiovisual content. The oracle selective grounding demonstrates an upper bound; future work should develop automatic quality‑estimation models that can identify low‑quality baseline segments without reference translations, enabling practical selective grounding. Human evaluation of pragmatic adequacy (e.g., honorifics, emotional tone) would complement automatic metrics to better capture the subtle benefits of selective visual grounding. Additionally, exploring more sophisticated alignment techniques (e.g., dynamic time warping) could further reduce temporal misalignment and improve the robustness of fine‑grained visual context.

\section*{Acknowledgements}
The authors would like to express their sincere gratitude to the project, Centre of Indian Language Data (COIL-D) under Bhashini, funded by the Ministry of Electronics and Information Technology (MeitY), Government of India for its generous support. 

\bibliography{eamt26}

@inproceedings{papineni-etal-2002-bleu,
    title = "{B}leu: a Method for Automatic Evaluation of Machine Translation",
    author = "Papineni, Kishore  and
      Roukos, Salim  and
      Ward, Todd  and
      Zhu, Wei-Jing",
    editor = "Isabelle, Pierre  and
      Charniak, Eugene  and
      Lin, Dekang",
    booktitle = "Proceedings of the 40th Annual Meeting of the Association for Computational Linguistics",
    month = jul,
    year = "2002",
    address = "Philadelphia, Pennsylvania, USA",
    publisher = "Association for Computational Linguistics",
    url = "https://aclanthology.org/P02-1040/",
    doi = "10.3115/1073083.1073135",
    pages = "311--318"
}

@inproceedings{singh-etal-2025-evaluation,
    title = "Evaluation of {LLM} for {E}nglish to {H}indi Legal Domain Machine Translation Systems",
    author = "Singh, Kshetrimayum Boynao  and
      Kumar, Deepak  and
      Ekbal, Asif",
    editor = "Haddow, Barry  and
      Kocmi, Tom  and
      Koehn, Philipp  and
      Monz, Christof",
    booktitle = "Proceedings of the Tenth Conference on Machine Translation",
    month = nov,
    year = "2025",
    address = "Suzhou, China",
    publisher = "Association for Computational Linguistics",
    url = "https://aclanthology.org/2025.wmt-1.57/",
    doi = "10.18653/v1/2025.wmt-1.57",
    pages = "823--833",
    ISBN = "979-8-89176-341-8"
}

@book{eberhard2025ethnologue,
  author = {Eberhard, David M. and Simons, Gary F. and Fennig, Charles D.},
  title = {Ethnologue: Languages of the World},
  year = {2025},
  edition = {28th},
  publisher = {SIL International},
  address = {Dallas, Texas},
  url = {http://www.ethnologue.com}
}

@inproceedings{wang-zhao-2024-tram,
    title = "{TRAM}: Benchmarking Temporal Reasoning for Large Language Models",
    author = "Wang, Yuqing  and
      Zhao, Yun",
    editor = "Ku, Lun-Wei  and
      Martins, Andre  and
      Srikumar, Vivek",
    booktitle = "Findings of the Association for Computational Linguistics: ACL 2024",
    month = aug,
    year = "2024",
    address = "Bangkok, Thailand",
    publisher = "Association for Computational Linguistics",
    url = "https://aclanthology.org/2024.findings-acl.382/",
    doi = "10.18653/v1/2024.findings-acl.382",
    pages = "6389--6415"
}

@inproceedings{elliott2016multi30k,
  title={Multi30k: Multilingual english-german image descriptions},
  author={Elliott, Desmond and Frank, Stella and Sima’an, Khalil and Specia, Lucia},
  booktitle={Proceedings of the 5th Workshop on Vision and Language},
  pages={70--74},
  year={2016}
}

@inproceedings{artetxe-etal-2020-cross-b,
    title = "On the Cross-lingual Transferability of Monolingual Representations",
    author = "Artetxe, Mikel  and
      Ruder, Sebastian  and
      Yogatama, Dani",
    editor = "Jurafsky, Dan  and
      Chai, Joyce  and
      Schluter, Natalie  and
      Tetreault, Joel",
    booktitle = "Proceedings of the 58th Annual Meeting of the Association for Computational Linguistics",
    month = jul,
    year = "2020",
    address = "Online",
    publisher = "Association for Computational Linguistics",
    url = "https://aclanthology.org/2020.acl-main.421/",
    doi = "10.18653/v1/2020.acl-main.421",
    pages = "4623--4637"
}

@misc{qwen2025qwen25technicalreport,
      title={Qwen2.5 Technical Report}, 
      author={Qwen and : and An Yang and Baosong Yang and Beichen Zhang and Binyuan Hui and Bo Zheng and Bowen Yu and Chengyuan Li and Dayiheng Liu and Fei Huang and Haoran Wei and Huan Lin and Jian Yang and Jianhong Tu and Jianwei Zhang and Jianxin Yang and Jiaxi Yang and Jingren Zhou and Junyang Lin and Kai Dang and Keming Lu and Keqin Bao and Kexin Yang and Le Yu and Mei Li and Mingfeng Xue and Pei Zhang and Qin Zhu and Rui Men and Runji Lin and Tianhao Li and Tianyi Tang and Tingyu Xia and Xingzhang Ren and Xuancheng Ren and Yang Fan and Yang Su and Yichang Zhang and Yu Wan and Yuqiong Liu and Zeyu Cui and Zhenru Zhang and Zihan Qiu},
      year={2025},
}

@inproceedings{popovic-2015-chrf,
    title = "chr{F}: character n-gram {F}-score for automatic {MT} evaluation",
    author = "Popovi{\'c}, Maja",
    editor = "Bojar, Ond{\v{r}}ej  and
      Chatterjee, Rajan  and
      Federmann, Christian  and
      Haddow, Barry  and
      Hokamp, Chris  and
      Huck, Matthias  and
      Logacheva, Varvara  and
      Pecina, Pavel",
    booktitle = "Proceedings of the Tenth Workshop on Statistical Machine Translation",
    month = sep,
    year = "2015",
    address = "Lisbon, Portugal",
    publisher = "Association for Computational Linguistics",
    url = "https://aclanthology.org/W15-3049/",
    doi = "10.18653/v1/W15-3049",
    pages = "392--395"
}

@inproceedings{rei-etal-2020-comet,
    title = "{COMET}: A Neural Framework for {MT} Evaluation",
    author = "Rei, Ricardo  and
      Stewart, Craig  and
      Farinha, Ana C  and
      Lavie, Alon",
    editor = "Webber, Bonnie  and
      Cohn, Trevor  and
      He, Yulan  and
      Liu, Yang",
    booktitle = "Proceedings of the 2020 Conference on Empirical Methods in Natural Language Processing (EMNLP)",
    month = nov,
    year = "2020",
    address = "Online",
    publisher = "Association for Computational Linguistics",
    url = "https://aclanthology.org/2020.emnlp-main.213/",
    doi = "10.18653/v1/2020.emnlp-main.213",
    pages = "2685--2702"
}

@misc{vasu2025fastvlmefficientvisionencoding,
      title={FastVLM: Efficient Vision Encoding for Vision Language Models}, 
      author={Pavan Kumar Anasosalu Vasu and Fartash Faghri and Chun-Liang Li and Cem Koc and Nate True and Albert Antony and Gokul Santhanam and James Gabriel and Peter Grasch and Oncel Tuzel and Hadi Pouransari},
      year={2025},
      eprint={2412.13303},
      archivePrefix={arXiv},
      primaryClass={cs.CV},
      url={https://arxiv.org/abs/2412.13303}, 
}

@misc{llama3,
    title = {Llama 3.1: The Llama 3.1 collection of multilingual large language models},
    author = {{Meta}},
    year = {2024},
    month = {July},
    howpublished = {\url{https://huggingface.co/meta-llama/Llama-3.1-8B}},
    note = {Model release date: July 23, 2024. Accessed: 2026-03-27}
}

@article{10.1145/3748311,
author = {Gain, Baban and Bandyopadhyay, Dibyanayan and Mukherjee, Samrat and Adak, Chandranath and Ekbal, Asif},
title = {Impact of Visual Context on Noisy Multimodal NMT: An Empirical Study for English to Indian Languages},
year = {2025},
issue_date = {August 2025},
publisher = {Association for Computing Machinery},
address = {New York, NY, USA},
volume = {24},
number = {8},
issn = {2375-4699},
url = {https://doi.org/10.1145/3748311},
doi = {10.1145/3748311},
journal = {ACM Trans. Asian Low-Resour. Lang. Inf. Process.},
month = aug,
articleno = {79},
numpages = {27}
}

@incollection{radinger2025subtitling,
  title={Subtitling in Audiovisual Translation Studies},
  author={Radinger, Anke},
  booktitle={Researching Subtitling Processes},
  pages={29--53},
  year={2025},
  publisher={Springer}
}

@inproceedings{appicharla-etal-2024-case,
    title = "A Case Study on Context-Aware Neural Machine Translation with Multi-Task Learning",
    author = "Appicharla, Ramakrishna  and
      Gain, Baban  and
      Pal, Santanu  and
      Ekbal, Asif  and
      Bhattacharyya, Pushpak",
    booktitle = "Proceedings of the 25th Annual Conference of the European Association for Machine Translation (Volume 1)",
    month = jun,
    year = "2024",
    address = "Sheffield, UK",
    publisher = "European Association for Machine Translation (EAMT)",
    url = "https://aclanthology.org/2024.eamt-1.21/",
    pages = "246--257"
}

@inproceedings{singh-etal-2025-evaluating,
    title = "Evaluating {I}ndic{T}rans2 and {B}y{T}5 for {E}nglish{--}{S}antali Machine Translation Using the Ol Chiki Script",
    author = "Singh, Kshetrimayum Boynao  and
      Ekbal, Asif  and
      Pakray, Partha",
    editor = "Shukla, Ankita  and
      Kumar, Sandeep  and
      Bedi, Amrit Singh  and
      Chakraborty, Tanmoy",
    booktitle = "Proceedings of the 1st Workshop on Multimodal Models for Low-Resource Contexts and Social Impact (MMLoSo 2025)",
    month = dec,
    year = "2025",
    address = "Mumbai, India",
    publisher = "Association for Computational Linguistics",
    url = "https://aclanthology.org/2025.mmloso-1.9/",
    pages = "95--100",
    ISBN = "979-8-89176-311-1"
}

@inproceedings{kumar-etal-2025-vision,
    title = "Does Vision Still Help? Multimodal Translation with {CLIP}-Based Image Selection",
    author = "Kumar, Deepak  and
      Gain, Baban  and
      Singh, Kshetrimayum Boynao  and
      Ekbal, Asif",
    editor = "Nakazawa, Toshiaki  and
      Goto, Isao",
    booktitle = "Proceedings of the Twelfth Workshop on Asian Translation (WAT 2025)",
    month = dec,
    year = "2025",
    address = "Mumbai, India",
    publisher = "Association for Computational Linguistics",
    url = "https://aclanthology.org/2025.wat-1.12/",
    doi = "10.18653/v1/2025.wat-1.12",
    pages = "115--123",
    ISBN = "979-8-89176-309-8"
}

@inproceedings{datta-etal-2025-findings,
    title = "Findings of the {JUST}-{NLP} 2025 Shared Task on Summarization of {I}ndian Court Judgments",
    author = "Datta, Debtanu  and
      Paul, Shounak  and
      Singh, Kshetrimayum Boynao  and
      Kumar, Sandeep  and
      Joshi, Abhinav  and
      Mishra, Shivani  and
      Jain, Sarika  and
      Ekbal, Asif  and
      Goyal, Pawan  and
      Modi, Ashutosh  and
      Ghosh, Saptarshi",
    booktitle = "Proceedings of the 1st Workshop on NLP for Empowering Justice (JUST-NLP 2025)",
    month = dec,
    year = "2025",
    address = "Mumbai, India",
    publisher = "Association for Computational Linguistics",
    url = "https://aclanthology.org/2025.justnlp-main.2/",
    doi = "10.18653/v1/2025.justnlp-main.2",
    pages = "5--11",
    ISBN = "979-8-89176-312-8"
}

@inproceedings{shen-etal-2025-coe,
    title = "{C}o{E}: A Clue of Emotion Framework for Emotion Recognition in Conversations",
    author = "Shen, Zhiyu  and
      Pang, Yunhe  and
      Rao, Yanghui  and
      Yu, Jianxing",
    editor = "Che, Wanxiang  and
      Nabende, Joyce  and
      Shutova, Ekaterina  and
      Pilehvar, Mohammad Taher",
    booktitle = "Proceedings of the 63rd Annual Meeting of the Association for Computational Linguistics (Volume 1: Long Papers)",
    month = jul,
    year = "2025",
    address = "Vienna, Austria",
    publisher = "Association for Computational Linguistics",
    url = "https://aclanthology.org/2025.acl-long.1148/",
    doi = "10.18653/v1/2025.acl-long.1148",
    pages = "23548--23563",
    ISBN = "979-8-89176-251-0"
}
\bibliographystyle{eamt26}

\appendix
\section{Appendix}

\subsection{Ethical Considerations}
Multimodal systems may inherit and amplify biases present in movies, where visual cues can reinforce stereotypes. Temporal misalignment between audio and visuals may also lead to inappropriate translations of culturally sensitive content. For Indian languages, honorifics and regional variations require careful handling to ensure accuracy. We advocate for human oversight, transparent reporting, and evaluation frameworks that assess cultural appropriateness. As selective grounding shows that only some segments need visual context, applying visual enhancement through a bias-aware filter can help reduce potential harms.

\subsection{Visual Feature Extraction}
\begin{enumerate}
    \item The original \texttt{.mkv} video is compressed to $\approx$ 1-GB and converted to \texttt{.mp4}.
    \item Frames are extracted at 1-fps and passed to Apple FastVLM-0.5B to obtain a raw textual description per frame.
    \item Raw descriptions are cleaned (removing repeated phrases, normalising punctuation) to produce a final description per frame.
\end{enumerate}
For each subtitle, the visual context is formed by concatenating the cleaned descriptions of all frames whose timestamps fall within the subtitle’s time window. When multiple frames belong to the same segment, the descriptions are aggregated into a single summary (for the gap‑based method, the window is the interval between the previous subtitle’s end and the current subtitle’s start).

\subsection{Context-Aware Translation Pipeline}
The translation pipeline uses the same Qwen-2.5-7B-Instruct model for both baseline and visual‑enhanced translations. The only difference is the input prompt. For the baseline, the model receives only the English source. For visual‑enhanced, we prepend the summarised visual context (Attr‑VC or Inter-VS) to the source using the visual‑enhanced prompt template shown in Table~\ref{tab:prompts}. The prompt instructs the model to ground its translation in the visual scene, paying attention to gender, honorifics, and emotional tone. The same greedy decoding parameters described in hyperparameterss are used for all runs.

\subsection{Hyperparameters}
All inference runs use greedy decoding to ensure reproducibility.

For the translation model (Qwen-2.5-7B-Instruct):
\begin{itemize}[noitemsep]
    \item \texttt{max\_new\_tokens = 100}
    \item \texttt{do\_sample = False}
    \item \texttt{repetition\_penalty = 1.1}
    \item temperature and top‑p left at default (1.0 and 1.0, effectively greedy).
\end{itemize}

For the summarization model (Llama-3.1-8B-Instruct):
\begin{itemize}[noitemsep]
    \item \texttt{max\_new\_tokens = 256}
    \item \texttt{do\_sample = False}
    \item temperature and top‑p at default (1.0 and 1.0).
\end{itemize}

\subsection{Oracle Selective Grounding}
Per‑segment COMET scores for the baseline translation (against the reference) are computed using the \texttt{Unbabel/wmt22-comet-da} model. The worst \(k\%\) of segments (by baseline COMET) are replaced with the corresponding visual‑enhanced translation (from either Attr‑VC or Inter-VS). We report results for \(k = 20\%\) and \(30\%\). Appendix~\ref{tab:selective_appendix}

\section{Additional Analysis}

\subsection{Window Size Considerations}
The choice of a 5‑minute sliding window for Attribute Visual Context was guided by the need to capture enough local scene context while remaining robust to temporal drift. A larger window (e.g., 10 minutes) would aggregate more visual information, but it could also include more irrelevant frames, potentially increasing the risk of hallucination when the visual context does not align with the dialogue. A smaller window (e.g., 2 minutes) would be more sensitive to misalignment. The 5‑minute window offers a reasonable trade‑off, as evidenced by the improvement in COMET over the baseline for most language-movie pairs.

\subsection{Oracle Selective Grounding}
The full oracle selective results for both summarization methods are presented in Table~\ref{tab:selective_appendix}. Replacing only the worst 20-30\% of baseline segments consistently lifts COMET above the baseline, demonstrating that most of the gain can be achieved with a fraction of the visual processing.

\section{Additional Analysis}
The 5‑minute sliding window offers a reasonable trade‑off between context and robustness; a larger window would risk including irrelevant frames, a smaller window would be more sensitive to misalignment. Oracle selective grounding (Table~\ref{tab:selective_appendix}) shows that replacing only the worst 20-30\% of baseline segments recovers most of the gain with minimal visual processing. Fine‑grained fusion methods (visual prefixing and cross‑attention fusion) failed under misalignment (e.g., “He is very kind” → “He drives fast”) because they assume perfect alignment, whereas coarse attribute summarization ignores irrelevant frames. This reinforces the idea that alignment quality often outweighs architectural complexity, and that selective grounding offers a practical path to efficient visual‑guided translation.

\subsection{Clarifications}
Movie subtitle translation is inherently difficult (fragmented speech, cultural references, zero‑shot domain adaptation); our baseline scores reflect this challenge, and our contribution lies in the \emph{relative} COMET gain, not absolute SOTA. We use Qwen‑2.5‑7B‑Instruct because it supports zero‑shot instruction following without fine‑tuning, unlike dedicated Indic models (e.g., IndicTrans2) that are optimised for sentence‑level translation and do not accept multi‑field visual context prompts. FastVLM-0.5B prioritises efficiency (85× faster than LLaVA); using a smaller VLM makes gains harder to achieve, so our positive results demonstrate robustness. FastVLM outputs raw descriptions; we summarise them with Llama 3.1 to obtain structured attributes or free‑text gap summaries, because direct prompting of FastVLM for structured output is not feasible. Our experiments compare visual-enhanced translation against an identical text‑only baseline, isolating the effect of visual context; comparing across different MT systems would confound architectural differences.

\subsection{Evaluation Metrics}
We report corpus‑level BLEU~\cite{papineni-etal-2002-bleu} using SacreBLEU and chrF++~\cite{popovic-2015-chrf} with \texttt{word\_order=2}. COMET~\cite{rei-etal-2020-comet} is computed with the \texttt{wmt22-comet-da} model using default settings.

\subsection{Data and Code Availability}
To foster reproducibility, the curated movie-subtitle-visual alignment data for all five languages will be released under a fair‑use educational/research license. The release includes English sources, reference translations, and extracted visual descriptions. All the codes and datasets used for extraction, summarization, translation, and evaluation are available at GitHub \footnote{\url{https://github.com/Tarunc224/visually-guided-subtitle-translation}} and our group page.\footnote{https://ai-nlp-ml.github.io/resources.html}

\begin{table*}[t]
\centering
\small
\setlength{\tabcolsep}{6pt}
\begin{tabular}{p{0.95\textwidth}}
\toprule
\centering \textbf{Baseline Prompt (text-only)} \tabularnewline
\midrule
\begin{minipage}[t]{\linewidth}
\begin{verbatim}
<|im_start|>system
You are a translation expert. Translate dialogue from English to {target_language}.
RULES:
- Provide ONLY the translated {target_language} dialogue.
- DO NOT include explanations, or English text.
<|im_end|>
<|im_start|>user
[SOURCE]: "{row['english_dialogue']}"
[TASK]: Translate to {target_language} dialogue.
<|im_end|>
<|im_start|>assistant
\end{verbatim}
\end{minipage}
\\
\tabularnewline
\midrule
\centering \textbf{Visual-Enhanced Prompt} \tabularnewline
\midrule
\begin{minipage}[t]{\linewidth}
\begin{verbatim}
<|im_start|>system
You are a cinematic multimodal translator specializing in English-to-{target_language}.
Your goal is to provide a "grounded translation" where the choice of words depends on the visual scene.

RULES:
1. GENDER: Use the Visual Context to identify speaker/listener gender.
2. HONORIFICS: Determine social hierarchy from the scene (Formal vs. Informal).
3. LOOSE MEANING: Prioritize emotional intent and natural {target_language} flow.
4. Output ONLY the translated {target_language} dialogue text. No names, no English.
<|im_end|>
<|im_start|>user
[VISUAL CONTEXT]: {row['visual_context']}
[ENGLISH SOURCE]: "{row['english_source']}"
[TASK]: Based on the visual scene, provide the most natural {target_language} translation.
<|im_end|>
<|im_start|>assistant
\end{verbatim}
\end{minipage}
\\
\tabularnewline
\bottomrule
\end{tabular}
\caption{Comparison of baseline and visual-enhanced prompts.}
\label{tab:prompts}
\end{table*}

\begin{table*}[t]
\centering
\small
\begin{tabular}{p{0.23\textwidth} p{0.72\textwidth}}
\toprule
\textbf{Summarisation Method} & \textbf{Prompt Template} \\
\midrule
\emph{Attribute Visual Context (Attr‑VC)} & 
\begin{minipage}[t]{\linewidth}
\begin{verbatim}
<|begin_of_text|><|start_header_id|>system<|end_header_id|>
Identify these cinematic attributes to guide {target_language} translation:
[SETTING]: (e.g., Formal, Public, Intimate)
[GENDER]: (Speaker/Listener gender)
[RELATION]: (e.g., Stranger, Family, Hostile)
[HONORIFIC]: (language‑specific, e.g., APNI/TUMI for Bengali)
[SUMMARY]: (One sentence factual summary with emotional intent)
Output ONLY these tags.<|eot_id|><|start_header_id|>user<|end_header_id|>
Visual Data: {sample[:3000]}<|eot_id|>
<|start_header_id|>assistant<|end_header_id|>
\end{verbatim}
\end{minipage} \\
\midrule
\emph{Inter‑Chunk Visual Summarisation (Inter-VS)} &
\begin{minipage}[t]{\linewidth}
\begin{verbatim}
<|begin_of_text|><|start_header_id|>system<|end_header_id|>
You are a movie analyzer. Summarize the following visual descriptions
from {start_sec}s to {end_sec}s of the movie into 2-3 sentences.
Focus ONLY on the current location and character actions.
Do not use introductory filler.
<|eot_id|><|start_header_id|>user<|end_header_id|>
Visual Data: {text_blob[:2500]}
<|eot_id|><|start_header_id|>assistant<|end_header_id|>
\end{verbatim}
\end{minipage} \\
\bottomrule
\end{tabular}
\caption{Prompt templates used for visual summarization. Both the prompts are given to Llama-3.1-8B-Instruct. The attribute prompt outputs structured tags; the inter‑chunk prompt outputs a free‑text sentence. Placeholders in braces are replaced with actual data at runtime.}
\label{tab:Summarisation_prompts}
\end{table*}

\begin{table*}[t]
\centering
\small
\resizebox{\textwidth}{!}{%
\begin{tabular}{l p{2.5cm} p{2.5cm} p{2.5cm} p{2.5cm} p{2.7cm}}
\toprule
\textbf{Scenario} & \textbf{Source} & \textbf{Baseline (literal)} & \textbf{Visual‑Enhanced} & \textbf{Reference} & \textbf{Explanation} \\
\midrule
\multicolumn{6}{c}{\textit{Visual context helps}} \\
\midrule
Emotion & “I’m so sorry.” & “I am sorry.” & “I am truly sorry.” & “I am truly sorry.” & Facial expression conveys deeper remorse, captured by visual summary. \\
Action & “He’s coming!” & “He is coming.” & “The man is coming!” & “The man is coming!” & Visual identifies gender and urgency. \\
Honorific & “Please sit.” & “Sit.” & “Please sit (formal).” & “Please sit (formal).” & Formal setting triggers correct honorific. \\
\midrule
\multicolumn{6}{c}{\textit{Visual context backfires (temporal misalignment)}} \\
\midrule
Misalign & “He is very kind.” & “He is very kind.” & “He drives fast.” & “He is very kind.” & Car chase frame (drift) causes hallucination. \\
Misalign & “I’m flying.” & “I’m flying.” & “I’m swimming.” & “I’m flying.” & Calm ocean scene (drift) misleads. \\
\midrule
\multicolumn{6}{c}{\textit{Visual context backfires (irrelevant visuals)}} \\
\midrule
Neutral & “How was your day?” & “How was your day?” & “How was your day?” & “How was your day?” & No improvement; adds processing overhead. \\
\bottomrule
\end{tabular}
}
\caption{When visual context helps (emotion, action, honorifics) and when it backfires (temporal misalignment, irrelevant visuals). Visual‑enhanced outputs are from the better of the two summarization methods for each case. The misalignment examples are drawn from actual data where cumulative drift paired a kind remark with a car chase, and a flying statement with a calm ocean scene.}
\label{tab:visual_help_backfire}
\end{table*}

\begin{table*}[t]
\centering
\small
\resizebox{\textwidth}{!}{%
\begin{tabular}{l l c|c c|c c}
\toprule
\multirow{2}{*}{Movie} & \multirow{2}{*}{Language} & \multirow{2}{*}{Baseline COMET} & \multicolumn{2}{c|}{5‑Minute Slide Visual Attribute} & \multicolumn{2}{c}{Inter‑Chunk Visual Summarisation} \\
\cmidrule(lr){4-5} \cmidrule(lr){6-7}
 & & & Oracle selective 20\% & Oracle selective 30\% & Oracle selective 20\% & Oracle selective 30\% \\
\midrule
Avatar & Bengali  & 0.6298 & 0.6682 & 0.6829 & 0.6709 & 0.6865 \\
Avatar & Telugu  & 0.5257 & 0.5354 & 0.5390 & 0.5361 & 0.5390 \\
Avatar & Tamil  & 0.5352 & 0.5580 & 0.5650 & 0.5569 & 0.5639 \\
Avatar & Kannada & 0.4857 & 0.4933 & 0.4943 & 0.4928 & 0.4946 \\
\midrule
Oppenhimer & Bengali  & 0.7026 & 0.7224 & 0.7248 & 0.7210 & 0.7237 \\
Oppenhimer & Hindi  & 0.6467 & 0.6617 & 0.6642 & 0.6643 & 0.6690 \\
Oppenhimer & Telugu  & 0.5475 & 0.5604 & 0.5654 & 0.5600 & 0.5647 \\
Oppenhimer & Tamil  & 0.5366 & 0.5630 & 0.5715 & 0.5639 & 0.5715 \\
Oppenhimer & Kannada  & 0.4938 & 0.5066 & 0.5096 & 0.5065 & 0.5090 \\
\midrule
Skyfall & Bengali  & 0.6914 & 0.7064 & 0.7098 & 0.7044 & 0.7056 \\
Skyfall & Hindi  & 0.6026 & 0.6223 & 0.6258 & 0.6216 & 0.6245 \\
Skyfall & Telugu  & 0.5288 & 0.5430 & 0.5478 & 0.5415 & 0.5454 \\
Skyfall & Tamil  & 0.5350 & 0.5581 & 0.5659 & 0.5561 & 0.5639 \\
Skyfall & Kannada  & 0.4920 & 0.5013 & 0.5038 & 0.5014 & 0.5038 \\
\midrule
Spider 2& Bengali  & 0.7190 & 0.7337 & 0.7359 & 0.7305 & 0.7350 \\
Spider 2& Hindi  & 0.6459 & 0.6684 & 0.6746 & 0.6717 & 0.6786 \\
Spider 2& Telugu  & 0.5407 & 0.5527 & 0.5567 & 0.5527 & 0.5569 \\
Spider 2& Tamil  & 0.5448 & 0.5684 & 0.5753 & 0.5700 & 0.5761 \\
\midrule
Titanic & Bengali  & 0.6960 & 0.7114 & 0.7130 & 0.7120 & 0.7150 \\
Titanic & Hindi  & 0.6152 & 0.6293 & 0.6321 & 0.6320 & 0.6367 \\
Titanic & Telugu  & 0.5350 & 0.5456 & 0.5481 & 0.5465 & 0.5494 \\
Titanic & Kannada  & 0.4950 & 0.5037 & 0.5065 & 0.5049 & 0.5063 \\
\bottomrule
\end{tabular}
}
\caption{Oracle selective COMET scores for the two summarization methods. “Oracle selective 20\%” and “Oracle selective 30\%” replace the worst 20\% and 30\% of baseline segments (by baseline COMET) with the corresponding visual‑enhanced translation. The full visual‑enhanced results are reported in Table~\ref{tab:combined_results}.}
\label{tab:selective_appendix}
\end{table*}

\end{document}